# VIDEO PROCESSING FOR BARYCENTER TRAJECTORY IDENTIFICATION IN DIVING

Stefano Frassinelli, Alessandro Niccolai and Riccardo E. Zich

Dipartimento di Energia, Politecnico di Milano, Milan, Italy

## ABSTRACT

*The aim of this paper is to show a procedure for identify the barycentre of a diver by means of video processing. This procedure is aimed to introduce quantitative analysis tools and diving performance measurement and therefore in diving training. Sport performance analysis is a trend that is growing exponentially for all level athletes: it has been applied extensively in some sports such as cycling. Sport performance analysis has been applied mainly for high level athletes; in order to be used also for middle or low level athletes the proposed technique has to be flexible and low cost. Video processing is suitable to fulfil both these requirements. In diving, the first analysis that has to be done is the barycentre trajectory tracking.*

## 1. INTRODUCTION

Sports have been recently changed with the application of modern techniques [1]: the training techniques are changing and the evaluation during competitions is going to be integrated as much as possible with computational technique [2] in order to have a more precise and objective performance evaluation.

This trend is associate to the increase of sports popularity [3][4] and the associated increase of elder people that are practicing sports [5][6].

In this sort of sports euphoria, performance analysis is becoming every year much more important both for increasing the performances and to reduce injuries caused by wrong training. Athletes of all levels are involved in analysis of their performances. This is giving to many different sport a significant push toward higher and higher levels of competitions [7][8].

Diving is one of the sports involved in this trend. Even if the popularity of this sport is not at the levels of sports such as basket, football and baseball, analysing diving is interesting due to the fact that there is a large space for improving the performances, the trainings and the competition judgments.

The performances of diving are quite difficult to be objectively and engineeringly measured, but this is really important both for the athlete that can have an understanding of the efficiency of their training, both for competition organizers that can have a tool to make the judgment more objective.

In order to address the problem of diving performance measurement, some wearable devices have been introduced in literature [9]. This approach can give satisfactory results but it is affected by some important limitations: first of all, it can be used only during training because they are not allowed during competitions. Moreover, the data collection can be limited by the availability of athletes that are using these devices: in fact, it is not possible to gather data from other sources.

In order to solve this problem, video analysis has been here used. In other sports, like soccer [10], video analysis have been introduced some years ago and many generic techniques or ad-hoc procedures [11] have been implemented. Some studies have been done also for diving [12], but the techniques proposed in literature are quite expansive and consequently they are not suitable for many low-level athlete applications.

Video processing is powerful because it is a flexible technique and it is possible to gather data from all the video acquired, also when these have been recorded for other purposes, e.g. it is possible to use the Olympic Games video to have a comparison and thus to help the athlete to improve his diving technique.

In this paper video processing has been applied to track the trajectory of the diver. Barycentre trajectory analysis is important in diving because it has some direct consequence in the performance and some indirect consequence.

The metrics related with barycentre position that have been here studied are:

- Barycentre trajectory during the taking off and the dive: this is an important metric [13] to understand possible mistakes made by the athlete during the taking off, like lateral movements induced by a non-symmetric jump on the springboard [14], [15] or on the platform, or to detect an overall not effectivity of the taking off phase

- Barycentre position during the entrance in the water: this is a metric that is important because it is one of the evaluation parameters that are commonly used in competitions;

- Maximum barycentre height[16]: this parameter is important because it gives an information of the time and space at disposal of the athlete for making all the figures that are required for the jump[17], [18].

In order to fully understand the procedure proposed in this paper, it is required to understand the needs and the requirements that have driven the design phase.Need and requirement analysis is a fundamental step in technique designing [19]. This is useful to frame the rage of applications and to give a reasonable direction to the design phase. Moreover, it is important to finally assess the performance of the designed product, in this case the performance analysis technique.

The first, most important need is the flexibility of the technique: it should be used for almost any kind of video, for 10m platform diving and for 3m springboard diving. In these two conditions the recording conditions are different: in fact, while for 3m springboard it is possible to have a fixed camera, for the 10m it is not possible to have a fixed camera with enough resolution; moreover, in this case the available space can be a strict constrain.

The second need, is the requirement of a technique that can be applied also for non-expert diver, so it must be cheap. It is not possible to deal witha prohibitive cost equipment. This means also that, not having a specific equipment,it is usually possible to have vibrations in the video that have to be correctly processed.

The paper is structured as follows: in Section 2 an overview of the overall video process will be given. In Section 3 describes the mosaicking procedure, Section 4 the barycentre identification process and Section 5 the athlete tracking system will be analysed. In Section 6some brief conclusions will be drawn.

## 2. OVERVIEW OF THE VIDEO PROCESSING

Video analysis is an effective technique in diving analysis [12] because it does not require any additional device on the body of the divers that can influence performance (the psychological approach is really important especially during high level competitions [14]) or that are forbidden[20].

It is important that the proposed technique is enough flexible to be applied both in training and during competitions: in this way, it is possible to compare the response of the athlete to the stress and also to have a first data to generate an expert system that is able to reproduce the judgment of the jury.

Another important advantage of video processing is that it can be applied to official video of competitions (Olympic Games or other International Competitions). These videos can be used as benchmark in the training.

Moreover, video processing results can be combined with kinematic simulations [21] to give to trainers and to athlete a complete tool in training that exploit the concept of *digital twin* that is becoming a paradigm for mechanical process analysis mainly in aerospace industry[22].

The overall process is described in Figure 1. It is composed by four steps and it is aimed to pass from a video, acquired during training or found on the Internet, to a barycentre trajectory.

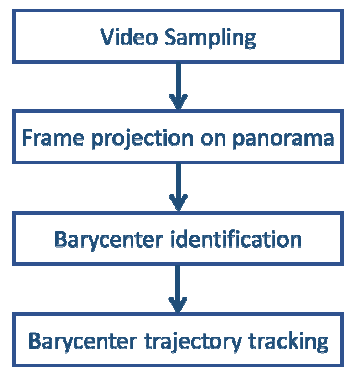

Figure 1: process flow chart

The sampling of the video should be done considering a trade-off between the accuracy and the computational time required for the analysis. There is an important constraint related with this step: aliasing should be avoided.

In order to avoid aliasing, it is necessary to have a high frequency rate. Usually a dive from a 10m platform lasts between 2 and 3 seconds. During the dive, athletes have to perform at most 6 figures, so the minimum frequency required to avoid aliasing is 6 Hz.

Increasing the sampling frequency means having more data to reject noise or errors in the analysis; on the other hand, it means a higher computational time. A good trade-off between these two aspects is to set the sampling frequency to 20 frame per second.

In this way, the analysis of a dive corresponds to the analysis of approximately 60 frames. This can be acceptable because the computational capability of modern PC is enough for this purpose.

After having done a correct sampling from the video, it is possible to analyse firstly all the frames with the mosaicking procedure aimed to get the *panorama* picture that correspond to a picture in which all the frames have been stitched exploit common invariant features. Having done the panorama, it is possible to project all the frames one by one on the panorama: the output is a sequence of frames that contains all the background and the athlete in the right position of each frame.

It is possible to understand that doing this means having a set of frames that have the same information content with respect to a set of frames obtained with a fixed camera record that is able to get all the scene together: the advantages of the procedure with respect to the fixed camera are the possibility to have each frame with a higher resolution, the reduction of the required space for the recording and the possibility to use the same procedure to eliminate the vibrations.

The outcome of the previous procedure can be used to identify the barycentre of the athlete with an appropriate filtering of the image. In this case, the filtering adopted is a colour thresholding in the HSV colour space. The details of this passage will be described in detail in Section4.

Finally, the last step of the procedure is the barycentre trajectory identification. This step is easy because it requires only a time domain filtering of the trajectory in order to reject all high frequency variations that can be induced only by numerical approximations.

## 3. MOSAICKING TECHNIQUE

Image mosaicking is a computational techniquethat exploit the presence of common features in a set of pictures[23] in images to a picture, called *panorama*[24]. It is possible to refer to this technique also with the name of *image stitching* [25]. The basic of the problem consists in the identification of the matrix that describe the holography from an image to another. It has been extensively applied in many research and industrial fields, such as medicine [26] or power plant inspection [27].

This technique is based on the identification of a set of features that are kept invariant during the most common transformations between pictures. The features of different pictures are matched and then the geometric transformation is identified. In this step to each picture a transfer function is associated [28].

At this point it is possible to wrap all the image in one image, called *panorama*. The *panorama*can be interpreted in this context as the common background to all the pictures.This step can solve two of the problems highlighted in the Introduction: the presence of vibrations and the camera following the diver.

Algorithm 1 shows the procedure applied for image mosaicking.

| |
|---|
| 1.  Definition of the reference frame |
| 2.  Individuation of SURF features of the reference frame |
| 3.  **For** all the frames **do** |
| 4.      Identification of SURF feature of the frame |
| 5.      Feature matching between current frame and previous one |
| 6.      Evaluation of the affine transformation between the two frames |
| 7.  **End For** |
| 8.  Definition of the size of the panorama |
| 9.  Empty panorama creation |
| 10. **For** all the frames **do** |
| 11.     Frame transformation in the global reference system |
| 12.     Transformed frame added to the panorama |
| 13. **End For** |

Algorithm 1.  Mosaicking procedure

The first point is the definition of the reference frame. This is an important choice only when dealing with a set of images characterized by a high translational component: the image wrapping technique is based on the inverse of transformation matrix and the translational components are off-diagonal terms. If they are too big, the matrix becomes close to be singular and thus the result may be affected by numerical errors and has to be handled with particular care.

The choice of the types of features to be identified is a key factor of the entire procedure: there are many different types of features that can be used [29] and the results can significantly change from a type of feature to another. There are two quality index that can be used to measure the performances of the feature extractor; they are: the quality of the identified transformation and the number of useful identified features. Since measuring the first index is hard, the second one has been used to drive the choice of the feature type. In order to make this choice, a comparative study has been done: five types of features extractors have been used on the same pairs of images and the number of matched points have been used as quality index. Figure 2 shows the results of this analysis.

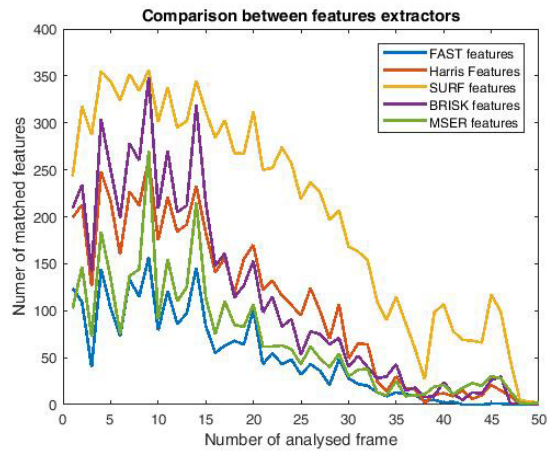

Figure 2.  Comparison between features extractors

The five types of features extracted are:

- FAST features, corner points introduced for high performance tracking of objects [30];

- Harris features, that are combined corners and edge detectors [31];

- SURF features, Speeded Up Robust Features [32];

- BRISK features, Binary Robust Invariant Scalable Keypoints [33];

- MSER features, Maximally Stable Extremal Regions [34];

From Figure 2 it is possible to see that the SURF feature extractor is the most performative in terms of number of matched features extracted.

The other kay choice in image mosaicking is the used types of transformations[24]: it has been seen that the affine transformation is the best one for this application [35] because allows enough degrees of freedom to the system but is able to reduce potential computational errors, while the perspective transformation is not able to do this, so it is possible to have important mismatching in the *panorama* creation.

After having created the *panorama,* it is possible to map each point of all the frames to the corresponding point of the *panorama*: in this way, it is possible to identify the barycentre in the frame and then to map it in the panorama.

It is possible to apply this technique to eliminate vibrations: in this case, shown in Figure 3, the difference between the background and each frame is low. This can be seen from Figure 3 because the black areas corresponding to pixel never present in the frames is small. This can be evaluated in a quantitative way extracting the translational components of all the transformations estimated while doing the mosaicking technique. The results of this extraction is shown in Figure 4.

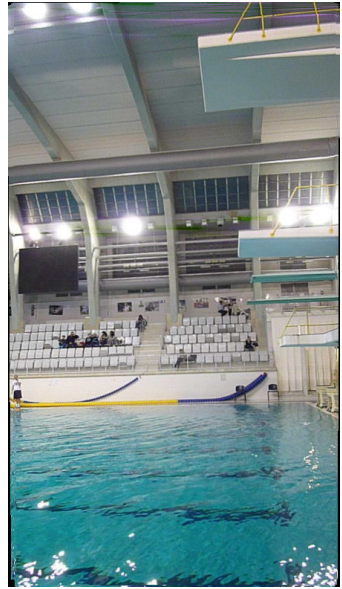

Figure 3.  Example of image mosaicking for vibration elimination

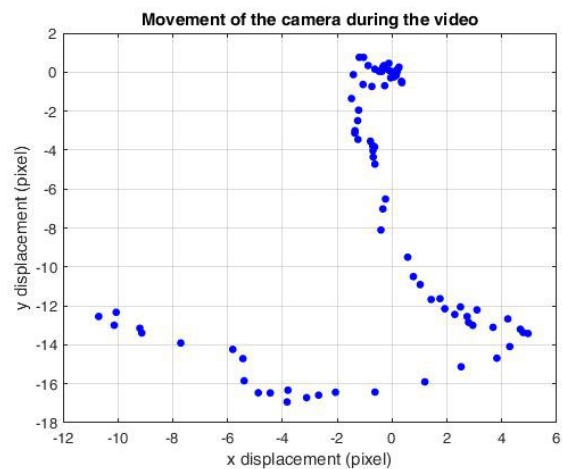

Figure 4.  Displacement of the camera during the non-professional video: the dots correspond to the position of the centre of the camera with respect to the centre of the reference frame

The same procedure can be applied on a video of a 10m platform dive of the Olympic Games of 2012 without changing the code. In this case study, the video has been recorded on order to have the athlete almost always at the centre of the video. The result of this technique applied to two different videos is shown in Figure 5. Also in this case has been possible to evaluate the displacement of the camera during the two videos. The results are shown in Figure 6.

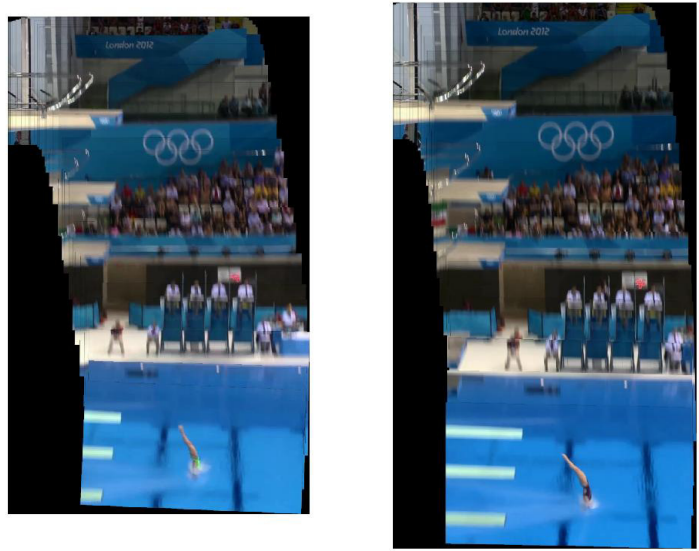

Figure 5. Two example of image mosaicking for background reconstruction with official Olympic Games video

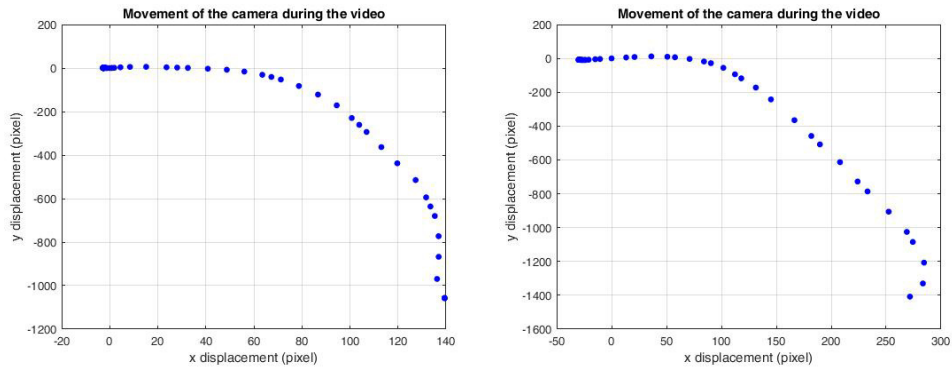

Figure 6. Movement of the camera with respect to the first frame reconstructed by the mosaicking procedure applied to the two videos of the Olympic Games

Comparing the results of Figure 4 and Figure 6, it is possible to see that the displacement of this last case is order of magnitude higher than the first one. This is obviously since the video is following an athlete that is diving from 10m platform; it has consequence on the results of the procedure because, as said before, highly translational transformations have quasi-singular matrixes.

## 4. BARYCENTRE IDENTIFICATION PROCEDURE

In this section, the barycentre identification procedure is described and some results of it are showed. Barycentre identification is a procedure that should be done frame by frame.

In this paper, this procedure has been implemented using a proper set of filters applied both on the frame and on the panorama. In this way, it is possible to find a reasonable compromise between the need to have a broadband filter and need to avoid large false positive recognitions that can bias the result. Algorithm 2 shows all the procedure steps.

1. Filter parameters setting
2. Panorama filtering
3. **For** all the frames **do**
4. Projection of the frame on the panorama
5. Frame colour filtering
6. Definition of the difference between the filtered panorama and the filtered frame
7. Object filtering
8. Barycentre calculation
9. **End For**

Algorithm 2. Barycentre identification procedure

It is possible to make some comments on the barycentre identification procedure:

- Due to the flexibility requirement, the colour filter [36] has to be not much selective because the light changes from frame to frame, so the colour of the diver can change. Obviously, a non-selective filter let many parts of the background in the filtered image. To reduce the number of false positive recognitions, also the background obtained from the mosaicking is filtered: in this way, the parts of the background that passes through the filter are known and they can be eliminated from the filtered frame;

- The colour filtering has been done by double threshold function applied to each channel in an appropriate colour space [37]. After several trials, the most suitable colour space is the HSV colour space [38].

Figure 7 shows two examples of the barycentre identification: the area output of the colour filter is the one with the white contour. The calculated barycentre is represented by the red square.

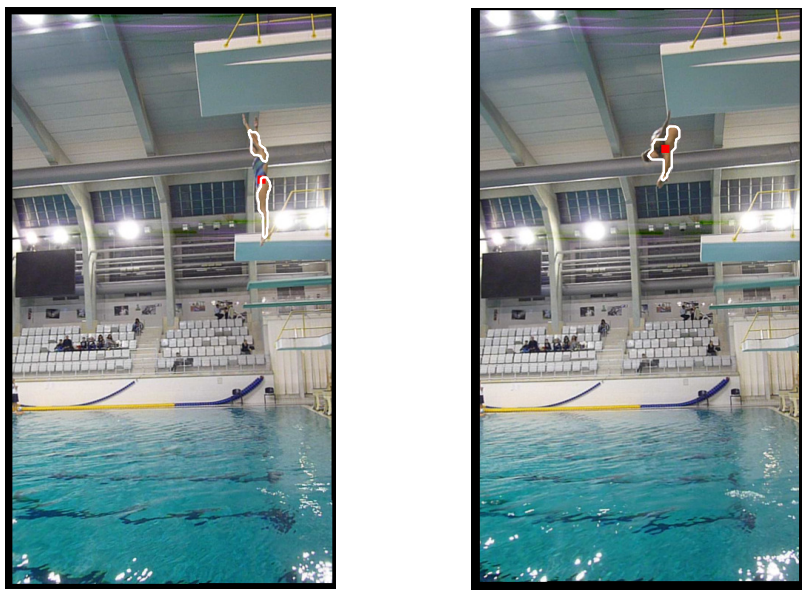

Figure 7. Example of barycentre identification in the non-professional video

Analysing Figure 7 it is possible to see that, even if the filter is not perfect due to the presence of the swimsuit and due to the darker illumination of the arms of the diver, the position estimated is a good approximation of the real position of the barycentre.

The same procedure has been also applied to the Olympic game video, as shown in Figure 8. This figure shows the filtered frame, the combination between the filtered frame and the filtered panorama and, then, the barycentre position.

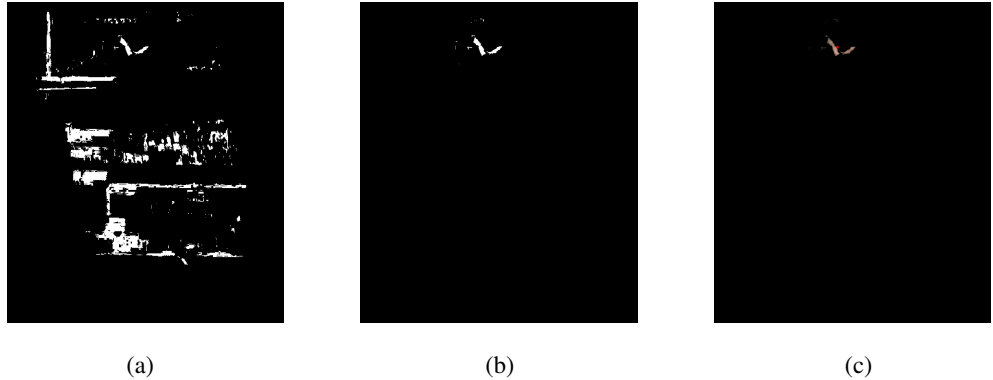

(a)                                (b)                                (c)

Figure 8.  Example of barycentre identification in the Olympic Games video: (a) shows the filtered frame (b) shows the filtered frame combined with the filtered background and (c) shows the barycentre position

## 5. BARYCENTRE POSITION TRACKING

The last step of the procedure is the barycentre tracking in all the frames and thus the reconstruction of the trajectory.

The issues present in this phase are related with the need to reject as much as possible the noise induced by mistakes in the barycentre identification without missing important information.

The procedure adopted to perform this step is a moving average filter applied to the reconstructed trajectory. The results of this procedure applied to the non-professional video are shown in Figure 9.

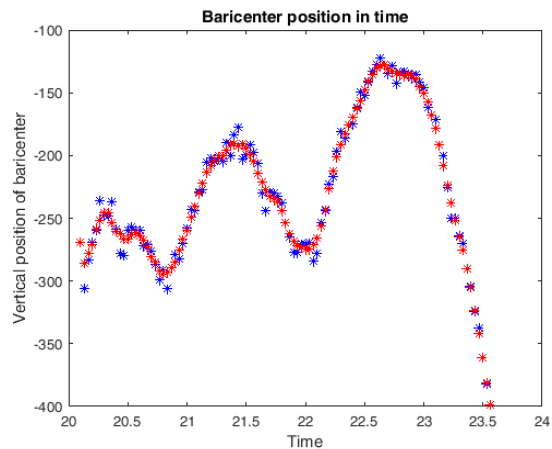

Figure 9.  Barycentre vertical position in time. The blue dots are the positions directly taken from the analysis, while the red dots are the results of a filtering procedure

Figure 9 shows the comparison between the original position and the filtered ones: it is possible to notice that the filtered signal is able to reproduce correctly the large-scale movement of the athlete without introducing a delay. Moreover, the noise present in the original signal is correctly rejected.

# 6. CONCLUSIONS

In this paper a simple, economic and flexible procedure of video analysis of diving performance assessment has been presented. The technique is based on several steps that make the system as flexible as possible.

A possible improvement of the technique is the use of Deep Learning Neural Networks. Even if this last method is really powerful, it requires a huge number of in-put test cases: it is possible to apply the procedure described in this paper to prepare these inputs for the Network training.